\documentclass{article}
\usepackage{spconf,amsmath,graphicx,enumitem}
\usepackage{diagbox}

\title{Evaluating Speech Synthesis by Training Recognizers on Synthetic Speech}
\name{Dareen Alharthi$^1$, Roshan Sharma$^1$, Hira Dhamyal$^1$, Soumi Maiti$^1$, Bhiksha Raj$^{1,2}$, Rita Singh$^1$}
\address{$^1$Carnegie Mellon University, $^2$Mohammed bin Zayed University of AI}

\usepackage[
backend=biber,
style=ieee,
citestyle=numeric-comp,
maxbibnames=10,
maxcitenames=10,
doi=false,isbn=false,url=false,eprint=false
]{biblatex}

\defbibheading{bibliography}[\refname]{}

\DeclareSourcemap{
	\maps[datatype=bibtex, overwrite=true]{
		\map{
			\step[fieldsource=booktitle,
			match=\regexp{.*Interspeech.*},
			replace={Proc. Interspeech}]
			\step[fieldsource=journal,
			match=\regexp{.*INTERSPEECH.*},
			replace={Proc. Interspeech}]
			\step[fieldsource=booktitle,
			match=\regexp{.*ICASSP.*},
			replace={Proc. ICASSP}]
			\step[fieldsource=booktitle,
			match=\regexp{.*icassp_inpress.*},
			replace={Proc. ICASSP (in press)}]
			\step[fieldsource=booktitle,
			match=\regexp{.*Acoustics,.*Speech.*and.*Signal.*Processing.*},
			replace={Proc. ICASSP}]
			\step[fieldsource=booktitle,
			match=\regexp{.*International.*Conference.*on.*Learning.*Representations.*},
			replace={Proc. ICLR}]
			\step[fieldsource=booktitle,
			match=\regexp{.*International.*Conference.*on.*Computational.*Linguistics.*},
			replace={Proc. COLING}]
			\step[fieldsource=booktitle,
			match=\regexp{.*SIGdial.*Meeting.*on.*Discourse.*and.*Dialogue.*},
			replace={Proc. SIGDIAL}]
			\step[fieldsource=booktitle,
			match=\regexp{.*International.*Conference.*on.*Machine.*Learning.*},
			replace={Proc. ICML}]
			\step[fieldsource=booktitle,
			match=\regexp{.*North.*American.*Chapter.*of.*the.*Association.*for.*Computational.*Linguistics:.*Human.*Language.*Technologies.*},
			replace={Proc. NAACL}]
			\step[fieldsource=booktitle,
			match=\regexp{.*Empirical.*Methods.*in.*Natural.*Language.*Processing.*},
			replace={Proc. EMNLP}]
			\step[fieldsource=booktitle,
			match=\regexp{.*Association.*for.*Computational.*Linguistics.*},
			replace={Proc. ACL}]
			\step[fieldsource=booktitle,
			match=\regexp{.*Automatic.*Speech.*Recognition.*and.*Understanding.*},
			replace={Proc. ASRU}]
			\step[fieldsource=booktitle,
			match=\regexp{.*Spoken.*Language.*Technology.*},
			replace={Proc. SLT}]
			\step[fieldsource=booktitle,
			match=\regexp{.*Speech.*Synthesis.*Workshop.*},
			replace={Proc. SSW}]
			\step[fieldsource=booktitle,
			match=\regexp{.*workshop.*on.*speech.*synthesis.*},
			replace={Proc. SSW}]
			\step[fieldsource=booktitle,
			match=\regexp{.*Advances.*in.*neural.*information.*processing.*},
			replace={Proc. NeurIPS}]
			\step[fieldsource=booktitle,
			match=\regexp{.*Advances.*in.*Neural.*Information.*Processing.*},
			replace={Proc. NeurIPS}]
			\step[fieldsource=booktitle,
			match=\regexp{.*Workshop.*on.* Applications.* of.* Signal.*Processing.*to.*Audio.*and.*Acoustics.*},
			replace={Proc. WASPAA}]
			\step[fieldsource=publisher,
			match=\regexp{.+},
			replace={{}}]
			\step[fieldsource=month,
			match=\regexp{.+},
			replace={{}}]
			\step[fieldsource=location,
			match=\regexp{.+},
			replace={{}}]
			\step[fieldsource=address,
			match=\regexp{.+},
			replace={{}}]
			\step[fieldsource=organization,
			match=\regexp{.+},
			replace={{}}]
		}
	}
}

\begin{document}
%
\maketitle
\begin{abstract}

Modern speech synthesis systems have improved significantly, with synthetic speech being indistinguishable from real speech. However, efficient and holistic evaluation of synthetic speech still remains a significant challenge. Human evaluation using Mean Opinion Score (MOS) is ideal, but inefficient due to high costs. Therefore, researchers have developed auxiliary automatic metrics like Word Error Rate (WER) to measure intelligibility. Prior works focus on evaluating synthetic speech based on pre-trained speech recognition models, however, this can be limiting since this approach primarily measures speech intelligibility. In this paper, we propose an evaluation technique involving the training of an ASR model on synthetic speech and assessing its performance on real speech. Our main assumption is that by training the ASR model on the synthetic speech, the WER on real speech reflects the similarity between distributions, a broader assessment of synthetic speech quality beyond intelligibility. Our proposed metric demonstrates a strong correlation with both MOS naturalness and MOS intelligibility when compared to SpeechLMScore and MOSNet on three recent Text-to-Speech (TTS) systems: MQTTS, StyleTTS, and YourTTS.








\end{abstract}
\section{Introduction}



Speech synthesis systems, aka Text-to-Speech (TTS) systems are increasingly becoming better. 
TTS systems are generally judged using the following two criteria: intelligibility and naturalness of the synthesized speech to human listeners. 
These metrics are traditionally measured by calculating  the Mean Opinion Score (MOS) (or intelligiblity score) of a panel of listeners, who annotate the synthesized speech with their subjective evaluation.
However, as is generally the norm for any annotation process involving human evaluators, computing MOS is time and resource-expensive. As an alternative, there are other proposed efficient algorithmically computable metrics \cite{cervnak2009diagnostic, maiti2023speechlmscore, sellam2023squid, le2023voicebox, binkowski2019high} which measure \textit{proxies} of intelligibility, quality and naturalness of the synthesized speech for example, such as using an ASR model trained on real speech to evaluate Word Error Rate (WER) of the synthesized speech. However, we argue that these metrics fall short of capturing the real quality of the synthesized speech. In this paper, we propose a better approximation of these measures of synthetic speech, which we show to be highly correlated with MOS. 

The top line of synthetic speech in intelligibility and naturalness is real speech, i.e. synthetic speech when closest to real speech would have the highest measure in these metrics. 
Therefore the question we pose in evaluating a TTS system is ``How close is the quality and intelligibility of synthetic speech generated by the system to that of real speech and how can we evaluate this?''.  We hypothesize that quality and intelligibility differences between the synthetic and real speech are attributable to the distributional shift between the two, and any metric which attempts to quantify these differences must capture this  shift. 
However explicit knowledge of the true distributions of the two is infeasible, and measurements must be made through mechanisms that invoke them implicitly.

Traditionally this is done by evaluating the synthetic speech on an ASR model trained on real speech. However, since speech synthesis is effectively a \textit{maximum likelihood} generating process that attempts to produce the most likely speech signal for any text, this can result in unrealistically high recognition accuracies biased in favor of the synthetic speech and, consequently, anomalous measurements of the speech quality. 
We argue that on the other hand, an ASR model trained on the synthetic speech and evaluated on real speech better captures the statistical difference between the two, and would be a better approximation of the closeness of the real and synthetic speech. Since the ASR models the distribution of the synthetic speech, its ability to recognize the real speech exhibits how closely the distributions of the synthetic training data matches with that of real testing data.

This paper makes the following contributions:
\begin{enumerate}[leftmargin=*,itemsep=0pt]
    \item We propose a new evaluation method for TTS that captures distributional similarity between real and synthetic speech as a proxy for perceptual speech quality tests. 
    \item We compare the proposed metric to multiple automatic metrics and Mean Opinion Score (MOS), and show that our metric correlates well with human-provided MOS.
\end{enumerate}
\label{sec:intro}

\section{Background: Speech Synthesis and Evaluation}
~\vspace{-0.1cm}
Recent advancements in speech synthesis systems have reached a point where they are often indistinguishable from human speech \cite{wang2023neural}. However, evaluating these systems has become increasingly complex. The most dependable method for evaluating speech synthesis systems from various perspectives is the Mean Opinion Score (MOS), in which human raters listen to synthesized speech and assess its naturalness, quality, and intelligibility using a 5-point Likert scale. However, this process is time-consuming, expensive, and subject to subjective judgments. To address these challenges, researchers have developed automatic metrics aimed at reducing evaluation costs. However, each metric is typically limited to evaluating a specific aspect of speech synthesis system performance, necessitating the use of multiple metrics to comprehensively assess these systems.
Recent studies have tackled this challenge through the training of regression models using pairs of speech MOS scores~\cite{lo2019mosnet} or by utilizing semi-supervised learning methods to acquire MOS scores. An important constraint associated with this method is the need for labeled datasets in the same domain, making it less generalizable~\cite{cooper2022generalization} to any text-to-speech (TTS) system.

Unsupervised metrics have also been employed to assess various aspects of speech synthesis, such as the Equal Error Rate for measuring speaker similarity in synthesized speech and metrics like Frechet DeepSpeech Distance~\cite{binkowski2019high} (FDSD) and Frechet Speech Distance (FSD)~\cite{le2023voicebox} to measure the quality and diversity of synthetic speech. However, it's important to note that each of these metrics focuses on a single factor and cannot serve as standalone measures. Recently, the utilization of speech-language models to assess speech quality has revealed a correlation with MOS scores. The SpeechLMScore~\cite{maiti2023speechlmscore} calculates the perplexity of synthetic speech by employing pretrained autoregressive unit speech language models (uLM)~\cite{lakhotia-etal-2021-generative}.
Another avenue of exploration involves Automatic Speech Recognition (ASR)-based metrics. One approach involves measuring the distance between synthetic and real speech \cite{minixhofer2022evaluating} by computing various distance metrics to assess speaker, prosody, and environmental similarity within real distributions. A commonly used ASR evaluation method is the computation of Word Error Rate (WER) \cite{cervnak2009diagnostic} for synthetic speech using pre-trained ASR models to measure intelligibility. Our proposed ASR evaluation approach seeks to evaluate both the naturalness and intelligibility of synthetic speech by quantifying the distribution shift between synthetic and real distributions.

\label{sec:format}

\section{Proposed Method}
\subsection{Divergence metric for Distributional Shift}

\begin{figure}
    \centering
    \includegraphics[width=0.45\columnwidth]{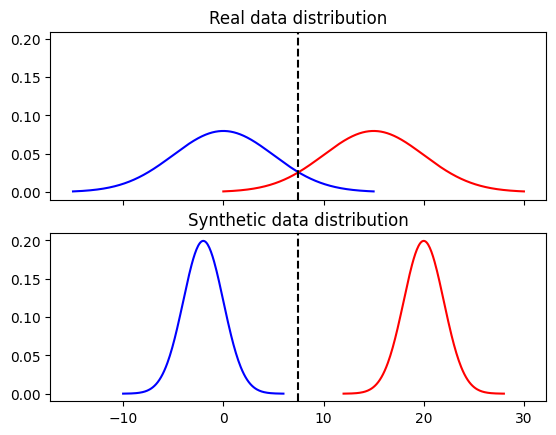}
    \includegraphics[width=0.45\columnwidth]{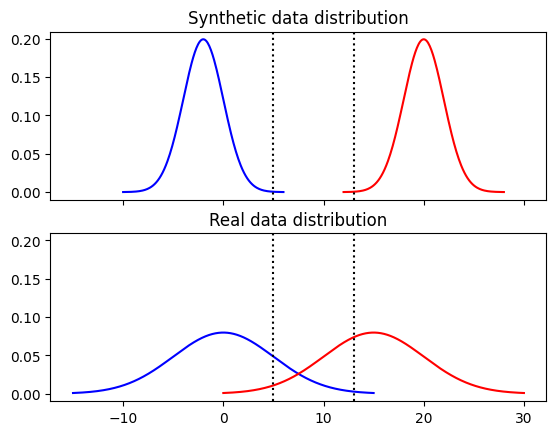}
    \caption{Left: Model trained on real data and tested on synthetic data.\\
    Right: Model trained on synthetic data and tested on real data.}
    \label{fig:real_synth_train_test}
\end{figure}

Given a text $T$, let $X_r$ be a random variable that represents real speech signals produced by humans to convey text $T$. Let $X_s$ be a random variable that represents synthesized speech from a TTS model for text $T$. $P(X_r, T)$ and $P(X_s, T)$ are the joint distributions of the speech and text. 

To evaluate the TTS Model, we want to compare these joint distributions - if the distributions are similar, synthetic speech has relatively high quality. Therefore, we want to compute a divergence $\text{div}(P(X_r,T), P(X_s,T))$ between the probability distributions, that measures the distance between the two distributions, i.e. distributional shift. 

Distributional shifts are typically computed using divergence metrics such as the Kulback-Leibler (KL) divergence~\cite{kullback1951information}, Jensen-Shannon divergence~\cite{lin1991jsd}, the Earth-mover distance~\cite{Rubner1998earthmovers} etc. However, these metrics require explicit knowledge of the distributions, or at least the ability to compute the probability of a given instance, if sampling-based approaches are to be used, which is infeasible, since it requires explicit modelling of $P(X_s, T)$ (or $P(X_r,T)$) whereas neural models only approximate the conditional probability of $T$.  Furthermore, even with explicit sampling, given the high dimensionality and time-series nature of the data, it would require sampling an infeasibly large number of  pairs of $(X, t)$, where $X$ is either $X_r$ or $X_s$, for reliable estimates.

To address the limitations of existing distributional similarity metrics, we propose an alternate metric that uses classification performances as a proxy to get distributional shifts. This classification-based pseudo-divergence uses probability distributions to get accuracy metrics. Given the two data distributions, input and labels, below we present the general case of the divergence metric.

Let $P_1$ and $P_2$ be two data distributions of random variables $X$ and $y$, where $X$ is the input signal and $y$ is the label. The predicted label for the given input $x$ can be written as: 
 $$\hat{y_1}(x) = \text{argmax}_y P_1(y|x)$$

When the classification boundaries are learned from $P_1$, and used to classify the data coming from the same distribution, the accuracy of this classification can be written as. 
$$ \mathbb{E}_{P_1(x)}[P_1(\hat{y}_1(x)|x)]$$

 Using the classification boundaries learned on the distribution $P_1$ and used to classify the data coming from the distribution $P_2$, the accuracy can be written as:
 $$\mathbb{E}_{P_2(x)}[P_2(\hat{y}_1(x)|x)]$$

The difference between the two classification accuracies captures the distributional shift between the $P_1$ and $P_2$. 
This can be written as:
$$ d(P_1, P_2) = |\mathbb{E}_{P_1(x)}[P_1(\hat{y}_1(x)|x)] - \mathbb{E}_{P_2(x)}[P_2(\hat{y}_1(x)|x)]|$$
The absolute value is needed since this difference could be negative. Note that $d(P_1, P_2)$ is a pseudo-divergence that goes to zero when $P_1 = P_2$ and is non-negative. It is also asymmetric, so $d(P_1, P_2) \neq d(P_2, P_1)$.

We can use the above formulation to calculate the pseudo-divergence of the real and synthetic speech. The distributions $P_1$ and $P_2$ can be estimated using an ASR Model trained on the data samples taken from the real and synthetic speech respectively. Since this is asymmetric, it is important to note which divergence to calculate $d(P_1, P_2)$ or $d(P_2, P_1)$. Either the ASR Model trained on real speech and tested on synthetic speech or vice versa. Empirically we show that the model trained on synthetic and tested on real data is a more accurate metric for the distributional shift of the two distributions than doing it vice versa. 
We explain the intuition behind this in the following section. 

\subsection{Real vs Synthetic data distribution}
Figure \ref{fig:real_synth_train_test} shows the joint distributions $X$ and $y$ where red curve shows when class $y=1$ and blue curve shows class $y=0$. Note that the real data has more variance than the synthetic data (which is true for the real and synthetic speech). 
When the classification boundary for the two classes is learned on the real data, there is some natural Bayes error associated with the class overlap present in the real data. When this classification boundary is used to do classification in the synthetic data, the error is zero, since the data distributions are far apart and there is no overlap in the two. In fact, there are multiple decision boundaries associated with different errors on the real data that would ensure zero error in the synthetic data. This zero error does not say anything about how different the real and synthetic data joint distributions are. The synthetic data distribution could be far off the chart, be highly unlikely compared to the real data, and still have zero error. 

On the other hand, let's consider the case where the lower variance data, i.e. the synthetic data, is used for learning the classification boundary. The right part of Figure \ref{fig:real_synth_train_test} shows this scenario. The dotted line shows the range where the decision boundary can lie such that the error rate on the synthetic data would be zero. However, this range of boundary would always be associated with greater than zero error on the real data. The higher the difference in the joint distributions in the real and synthetic distributions, the greater the range of errors in the real data. 

Therefore, the second scenario is better representative of the distributional differences in the real and synthetic data distributions. 
We believe that this would hold for the real and synthetic speech distributions. An ASR model trained on synthetic speech and evaluated on real speech would be a better metric of the quality of the synthetic speech than doing it the other way around.

\label{sec:pagestyle}

\section{Experimental setup}

\subsection{Text-to-Speech-Synthesis}

We evaluate the proposed method using three state-of-the-art open-source TTS systems: StyleTTS~\cite{li2022styletts}, MQTTS~\cite{chen2023vector}, and YourTTS~\cite{casanova2022yourtts}. These models utilize different techniques for synthesis, but all use a reference encoder to extract both speaker and style information from the input speech. For our assessments, we made use of the publicly released pre-trained models. The StyleTTS model, MQTTS, and YourTTS models we used were trained on LibriTTS~\cite{zen2019libritts}, Gigaspeech~\cite{chen21o_interspeech} without audiobooks, and VCTK~\cite{Yamagishi2019CSTRVC} respectively. A DeepPhonemizer~\cite{yolchuyeva2019transformer} was used to extract phonemes from the text for synthesis. 

\subsection{Automatic Speech Recognition}

In order to make evaluations robust and meaningful, we need to select strong End-to-End models. In this paper, we therefore elect to fine-tune Whisper rather than train from scratch using 10h of real/synthetic speech. We use the \texttt{Whisper-medium multilingual model}~\cite{radford2022robust} as the initialization. We then fine-tune it within ESPNet ~\cite{watanabe2018espnet,watanabeHybrid} using CTC loss ~\cite{graves2006connectionist}. ASR Inference was performed using beam search with a beam size of 5. 

\subsection{Datasets}
 To generate synthetic speech for our evaluation, we utilized the LibriTTS~\cite{zen2019libritts} dataset, which is based on Librispeech~\cite{panayotov2015librispeech}. From this dataset, we sample one subset of 10 hours
 containing speech data from all available speakers. All three TTS models used a speaker encoder to clone the identity of a given speech reference. It's worth noting that we excluded speech samples that were less than 4 seconds in duration and those exceeding 30 seconds in length. This exclusion was necessary as MQTTS and StyleTTS do not support short samples as references.

\subsection{Evaluation Metrics}

\noindent \textbf{MOS-Naturalness (MOS-N) }: We conducted a crowd-sourced Mean Opinion Score (MOS) evaluation to assess the naturalness of synthetic speech generated by each system, in comparison to real speech. We obtained 50 sentences from the LibriTTS test-clean dataset and another 50 from the LibriTTS test-other dataset, resulting in a total of 100 samples each for real speech, MQTTS, YourTTS and StyleTTS.  Each sample was evaluated by 10 raters, who were instructed to rate the naturalness of the speech on a scale of 1 to 5, with 1 indicating poor and 5 indicating excellent quality.\\

\noindent \textbf{MOS-Intelligibility(MOS-I)}: We assessed intelligibility of spoken words by using nonsense sentences \cite{kang2018empirical}, effectively eliminating sentence structure and grammar from the evaluation. This absence of structure allowed listeners to only focus on the quality of the synthesized speech and not be distracted by the grammar. 
Participants were presented with a choice between the original sentence and a transcription generated by the \texttt{Whisper-medium}. We specifically selected 60 sentences with relatively high Word Error Rate (WER) from a pool of 200 random sentences generated by ChatGPT~\cite{ouyang2022training}. Among these, 30 sentences were short (less than 10 words), while the other 30 were long. This allowed us to evaluate the impact of sentence length variation on intelligibility. We generated synthetic speech using the three TTS systems for the 60 sentences using a test-clean set as a reference for the model's speaker and style encoder. We used WebMushar \cite{schoeffler2015towards} to create a test form along with Prolific for crowd-sourcing.\\

\noindent \textbf{Intelligibility of Synthetic Speech using WER from Pre-trained ASR}: We computed the WER for synthetic speech generated by three different systems using the \texttt{Whisper medium multilingual}. This model is pre-trained on real speech and evaluated on synthetic speech. This setting of training / testing demonstrates the traditional way that speech synthesis evaluation is performed. 
This evaluation was performed on both the test-clean and test-other datasets from LibriTTS. 

\label{sec:typestyle}

\section{Experimental Results}

\title{Naturalness}
\begin{table}
  \caption{This table shows the scores for real and synthetic speech on multiple metrics for LibriTTS test-clean.  MOSNet and SpeechLMScore scores are on the same 50 samples of MOS-N. Relative ranking among synthetic speech systems are shown in red inside the brackets.}
  \centering 
\begin{threeparttable}







\resizebox{\columnwidth}{!}{%
\begin{tabular}{l|cccc} 
 \hline
  \diagbox{Metric}{Model} &  Ground Truth   & StyleTTS  &MQTTS& YourTTS\\
  \hline
  WER $\downarrow$ ~\cite{cervnak2009diagnostic} & 20.57 & \textbf{18.7}({\textcolor{red}{1}}) & 29.35({\textcolor{red}{3}}) & 22.1({\textcolor{red}{2}}) \\
 SpeechLMScore $\uparrow$ ~\cite{maiti2023speechlmscore}	 & 3.98 & 3.62 ({\textcolor{red}{3}})  & \textbf{4.13}({\textcolor{red}{1}}) & 3.96 ({\textcolor{red}{2}}) \\
 MOSNet  $\uparrow$ ~\cite{lo2019mosnet}	& 4.30 & \textbf{4.49}({\textcolor{red}{1}}) &  3.57({\textcolor{red}{3}})  & 4.01({\textcolor{red}{2}})  \\
  \hline
 MOS-N $\uparrow$	&\textbf{ 3.69}& 3.68 ({\textcolor{red}{1}})   & 3.66 ({\textcolor{red}{2}})  & 3.59 ({\textcolor{red}{3}}) \\
 MOS-I $\uparrow$	& - & \textbf{0.698}({\textcolor{red}{1}}) & 0.618({\textcolor{red}{2}})  & 0.566 ({\textcolor{red}{3}}) \\
  \hline
 Ours 10h $\downarrow$ &\textbf {3.1} & 3.3 ({\textcolor{red}{1}}) & 3.9({\textcolor{red}{2}})  & 4.5 ({\textcolor{red}{3}})   \\   

 \hline
\end{tabular}}
       
\end{threeparttable}
\label{tab:naturalness-comparison}
\end{table}

Table \ref{tab:naturalness-comparison} reports the results of our experiments on Libri-TTS with the proposed evaluation method. We consider multiple metrics and report raw scores of the metric in the rows and relative ranking scores in brackets next to the raw score. The first row, named WER shows the case when the model is trained on real data and evaluated on synthetic data. The last row shows our setting, where the model is trained on synthetic data and evaluated on real data.
Based on the absolute raw numbers of the metric, we rank the TTS systems from 1 to 3 based on which one performs the best to worst. For example in the row MOS-N, Style-TTS has the highest score and therefore has rank 1, followed by MQ-TTS and then YourTTS. 
In order to assess whether our metric is a good representation of the quality of synthetic speech, we compare the relative ranking of our metric with the other metrics. Two metrics with a matched relative ranking means that the metrics evaluate the quality of speech similarly and agree with each other.

First, we see that the Mean Opinion Score tests on naturalness (MOS-N) and intelligibility (MOS-I) agree on relative rankings between the synthetic speech models. Further, we observe that the traditionally used WER metric shown in the first row does not actually correlate completely with the MOS results. We observe similar issues with other popular metrics including SpeechLMScore and MOSNet. 

From the last row, we observe that our metric evaluation of synthetic speech has a similar trend as the reported MOS scores, matching both MOS-N and MOS-I. Compared with the inconsistent result from the first row and the consistent result from our metric, we demonstrate the importance of the proposed evaluation method. 









\label{sec:majhead}

\section{Conclusion}
In this paper, we address the challenge of automatic evaluation for synthetic speech by modeling the similarity/dissimilarity between the distributions of synthetic and real speech. Existing divergence metrics require a large number of samples to capture the joint distribution and hence it is infeasible to employ them to calculate the distributional shift. 
In this paper, we introduce a new divergence measure that can be computed without knowledge of the joint distribution. The metric uses an ASR model as an approximation for the data distributions and the WER as a proxy for the quality of the synthesized speech. The metric is asymmetric, and it matters what the speech recognition models are trained and tested on. We show that in practice it is more accurate to train the model on synthetic speech and assess the resulting model's performance on real speech than doing it vice versa. 
Experiments using 3 public open-source speech synthesis systems show that our model correlates positively with subjective human Mean Opinion Scores for naturalness and intelligibility, while previously used ways for evaluating ASR performance trained on real and evaluated on synthetic does not correlate. Further, we show that it only takes small amounts of synthetic speech to train the ASR model to be able to make reliable judgments on the quality of the synthesized speech. 
\label{sec:conclusion}

\pagebreak
\section{References}
\printbibliography


\end{document}